\documentclass[runningheads]{llncs}
\usepackage{graphicx}
\usepackage{enumitem}
\usepackage{amsmath,amssymb}
\usepackage{color}
\usepackage{mathrsfs}
\usepackage{multirow}
\usepackage[scriptsize]{subfigure}
\usepackage[linesnumbered,ruled]{algorithm2e}
\DeclareMathOperator{\F}{F}
\DeclareMathOperator{\T}{T}
\DeclareMathOperator{\tr}{tr}
\newcommand{\norm}[1]{\lVert#1\rVert}
\begin{document}
\title{Joint \& Progressive Learning from High-Dimensional Data for Multi-Label Classification}

\titlerunning{Joint \& Progressive Learning}
%
\author{Danfeng Hong\inst{1,2}\orcidID{0000-0002-3212-9584} \and
Naoto Yokoya\inst{3}\orcidID{0000-0002-7321-4590} \and
Jian Xu\inst{1}\orcidID{0000-0003-2348-125X} \and
Xiaoxiang Zhu\inst{1,2}\orcidID{0000-0001-5530-3613}}

\authorrunning{D. Hong et al.}
%

\institute{Remote Sensing Technology Institute (IMF), German Aerospace Center (DLR), Wessling, Germany\\
\email{\{danfeng.hong,jian.xu,xiao.zhu\}@dlr.de}
\and
Signal Processing in Earth Observation (SiPEO), Technical University of Munich, Munich, Germany\\
\and
RIKEN Center for Advanced Intelligence Project, Tokyo, Japan\\
\email{\{naoto.yokoya\}@riken.jp}}
\maketitle

\begin{abstract}
Despite the fact that nonlinear subspace learning techniques (e.g. manifold learning) have successfully applied to data representation, there is still room for improvement in explainability (explicit mapping), generalization (out-of-samples), and cost-effectiveness (linearization). To this end, a novel linearized subspace learning technique is developed in a joint and progressive way, called \textbf{j}oint and \textbf{p}rogressive \textbf{l}earning str\textbf{a}teg\textbf{y} (J-Play), with its application to multi-label classification. The J-Play learns high-level and semantically meaningful feature representation from high-dimensional data by 1) jointly performing multiple subspace learning and classification to find a latent subspace where samples are expected to be better classified; 2) progressively learning multi-coupled projections to linearly approach the optimal mapping bridging the original space with the most discriminative subspace; 3) locally embedding manifold structure in each learnable latent subspace. Extensive experiments are performed to demonstrate the superiority and effectiveness of the proposed method in comparison with previous state-of-the-art methods.

\end{abstract}
\section{Introduction}
\begin{figure*}[!t]
\label{Fig1}
\centering
        \includegraphics[width=10cm,height=8cm]{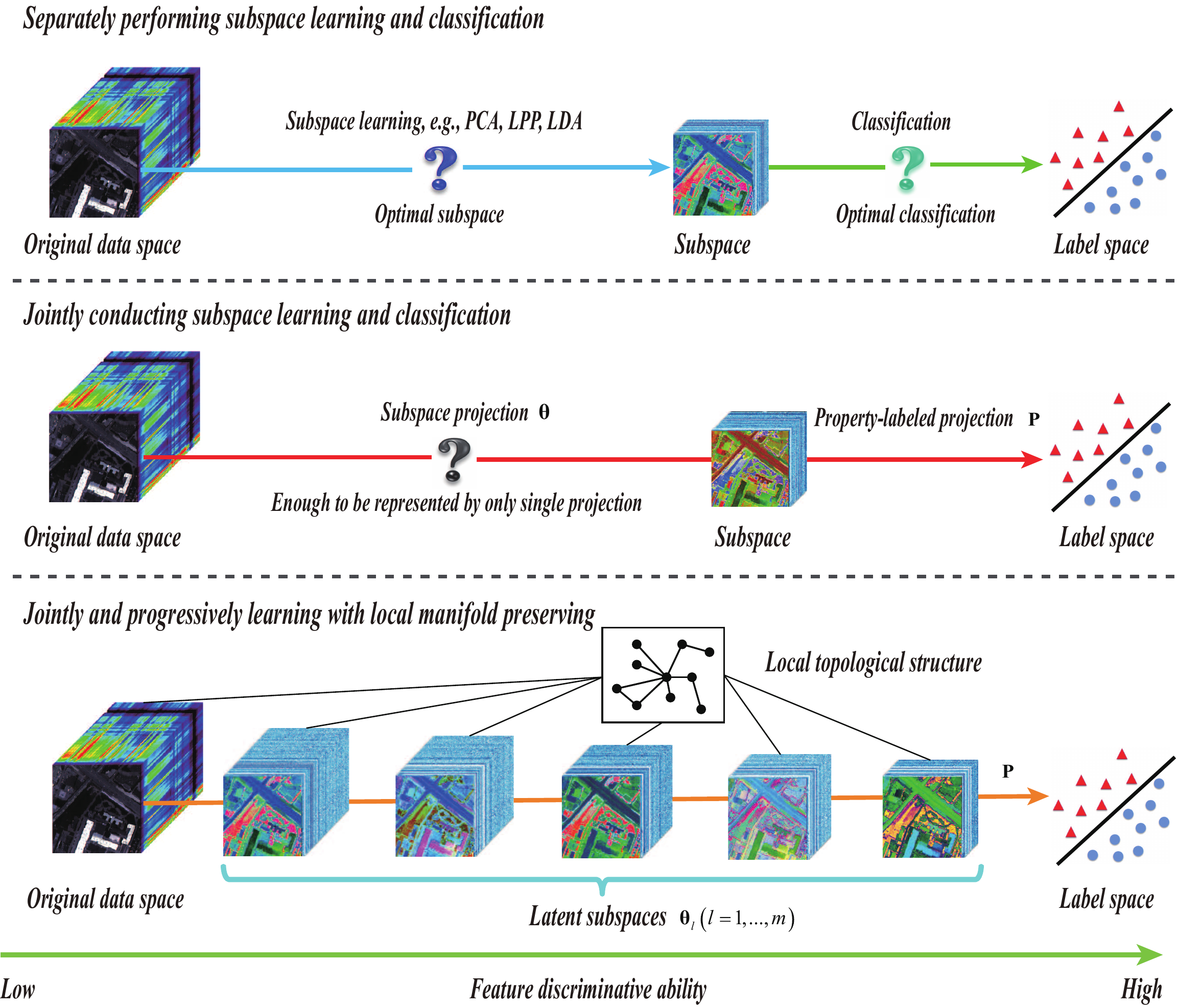}
\caption{The motivation interpolation from separately performing subspace learning and classification to joint learning to joint $\&$ progressive learning again. The subspaces learned from our model indicates the higher feature discriminative ability as explained by the green bottom line.}
\end{figure*}
High-dimensional data are often characterized by very rich and diverse information, which enables us to classify or recognize the targets more effectively and analyze data attributes more easily, but inevitably introduces some drawbacks (e.g. information redundancy, complex noise effects, high storage-consuming, etc.) due to \textit{the curve of dimensionality}. A general way to address this problem is to learn a low-dimensional and high-discriminative feature representation. In general, it is also called as dimensionality reduction or subspace learning. In the past decades, a large number of subspace learning techniques have been developed in the machine learning community, with successful applications to biometrics \cite{Martínez2001}\cite{He2005_2}\cite{hong2015novel}\cite{hong2016robust}, image/video analysis \cite{Tosato2010}, visualization \cite{Saul2003}, hyperspectral data analysis (e.g., dimensionality reduction and unmixing) \cite{hong2016local}\cite{Hong2017}\cite{hong2017learning}. These subspace learning techniques are generally categorized into linear or nonlinear methods. Theoretically, nonlinear approaches are capable of curving the data structure in a more effective way. There is, however, no explicit mapping function (poor explainability), and meanwhile it is relatively hard to embed the out-of-samples into the learned subspace (weak generalization) as well as high computational cost (lack of cost-effectiveness). Additionally, for a task of multi-label classification, these classic subspace learning techniques, such as principal component analysis (PCA) \cite{Wold1987}, local discriminant analysis (LDA) \cite{Martínez2001},  local fisher discriminant analysis (LFDA) \cite{Sugiyama2007}, manifold learning (e.g. Laplacian eigenmaps (LE) \cite{Belkin2003}, locally linear embedding (LLE) \cite{Roweis2000}) and their linearized methods (e.g. locality preserving projection (LPP)\cite{He2004}, neighborhood preserving embedding (NPE)\cite{He2005_1}), are commonly applied as a disjunct feature learning step before classification, whose limitation mainly lies in a weak connection between features by subspace learning and label space (see the top panel of Fig. 1). It is unknown which learned features (or subspace) can improve the classification.

Recently, a feasible solution to the above problems can be generalized as a joint learning framework \cite{ji2009linear} that simultaneously considers linearized subspace learning and classification, as illustrated in the middle panel of Fig. 1. Following it, more advanced methods have been proposed and applied in various fields, including supervised dimensionality reduction (e.g. least-squares dimensionality reduction (LSDR) \cite{Suzuki2013} and its variants: least-squares quadratic mutual information derivative (LSQMID) \cite{Tangkaratt2017}), multi-modal data matching and retrieval \cite{Wang2013,Wang2016}, and heterogeneous features learning for activity recognition \cite{Hu2015,Hu2016}. In these work, the learned features (or subspace) and label information are effectively connected by regression techniques (e.g. linear regression) to adaptively estimate a latent and discriminative subspace. Despite this, they still fail to find an optimal subspace, as single linear projection is hardly enough to represent the complex transformation from the original data space to the potential optimal subspace.

Motivated by the aforementioned studies, we propose a novel \textbf{j}oint and \textbf{p}rogre-ssive \textbf{l}earning str\textbf{a}teg\textbf{y} (J-Play) to linearly find an optimal subspace for general multi-label classification, illustrated in the bottom panel of Fig. 1. We practically extend the existing joint learning framework by learning a series of subspaces instead of single subspace, aiming at progressively converting the original data space to a potentially optimal subspace through multi-coupled intermediate transformations \cite{Kan2014}. Theoretically, by increasing the number of subspaces, coupled subspace variations are gradually narrowed down to a very small range that can be represented effectively via a \emph{linear transformation}. This renders us to find a good solution easier, especially when the model is complex and non-convex. We also contribute to structure learning in each latent subspace by locally embedding manifold structure.

The main highlights of our work can be summarized as follows:
\begin{itemize}[noitemsep,topsep=0pt]
\item A linearized progressive learning strategy is proposed to describe the variations from the original data space to potentially optimal subspace, tending to find a better solution. A joint learning framework that simultaneously estimates  subspace projections (connect the original space and the latent subspaces) and a property-labeled projection (connect the learned latent subspaces and label space) is considered to find a discriminative subspace where samples are expected to be better classified.
\item Structure learning with local manifold regularization is performed in each latent subspace.
\item Based on the above techniques, a novel joint and progressive learning strategy (J-Play) is developed for multi-label classification.
\item An iterative optimization algorithm based on the alternating direction method of multipliers (ADMM) is designed to solve the proposed model.
\end{itemize}
\section{Joint \& Progressive Learning Strategy (J-Play)}
\subsection{Notations}
Let $\mathbf{X}=\lbrack \mathbf{x}_{1},...,\mathbf{x}_{k},...,\mathbf{x}_{N}\rbrack \in\mathbb{R}^{d_{0}\times N}$ be a data matrix with $d_{0}$ dimensions and $N$ samples, and the matrix of corresponding class labels be $\mathbf{Y} \in\{0,1\}^{L \times N}$. The $k$th column of $\mathbf{Y}$ is $\mathbf{y}_{k}=\lbrack \mathbf{y}_{k1},...,\mathbf{y}_{kt},...,\mathbf{y}_{kL}\rbrack^{T}\in\mathbb{R}^{L\times 1}$ whose each element can be defined as follows:
\begin{equation}
\label{eq1}
  \mathbf{y}_{kt}=
    \begin{cases}
      \begin{aligned}
      1, \quad & \text{if \(\mathbf{y}_{k}\) belongs to the \(t\)-th class;}\\
      0, \quad & \text{otherwise.}
      \end{aligned}
    \end{cases}
\end{equation}
In our task, we aim to learn a set of coupled projections $\{\mathbf{\Theta}_{l}\}_{l=1}^{m}\in\mathbb{R}^{d_{l}\times d_{l-1}}$ and a property-labeled projection $\mathbf{P} \in\mathbb{R}^{L\times d_{m}}$, where $m$ stands for the number of subspace projections and $\{d_{l}\}_{l=1}^{m}$ are defined as the dimensions of those latent subspaces respectively, while $d_{0}$ is specified as the dimension of $\mathbf{X}$.
\subsection{Basic Framework of J-Play from the View of Subspace Learning}
Subspace learning is to find a low-dimensional space where we expect to maximize certain properties of the original data, e.g. variance (PCA), discriminative ability (LDA), and graph structure (manifold learning). Yan et al. \cite{Yan2007_2} summarized these subspace learning methods in a general graph embedding framework.

Given an undirected similarity graph $G=\{\mathbf{X},\mathbf{W}\}$ with the vertices $\mathbf{X}\in\{\mathbf{x}_{1},...,\mathbf{x}_{N}\}$ and the adjacency matrix $\mathbf{W}\in\mathbb{R}^{N\times N}$, we can intuitively measure the similarities among the data. By preserving the similarities relationship, the high-dimensional data can be well embedded into the low-dimensional space, which can be formulated by denoting the low-dimensional data representation as $\mathbf{Z}\in\mathbb{R}^{d\times N}$ ($d\ll d_{0}$) in the following
\begin{equation}
\label{eq2}
\begin{aligned}
       \mathop{\min}_{\mathbf{Z}}\tr(\mathbf{Z}\mathbf{L}\mathbf{Z}^{\T}), \quad \mathrm{s.t.} \quad \mathbf{Z}\mathbf{D}\mathbf{Z}^{\T}=\mathbf{I},
\end{aligned}
\end{equation}
where $\mathbf{D}_{ii}=\sum_{j}\mathbf{W}_{ij}$ is a diagonal matrix, $\mathbf{L}$ is a Laplacian matrix defined by $\mathbf{L}=\mathbf{D}-\mathbf{W}$ \cite{Chung1997}, and $\mathbf{I}$ is the identity matrix. In our case, we aim at learning multi-coupled linear projections to find optimal mapping, therefore a linearized subspace learning problem can be reformulated on the basis of Eq. (\ref{eq2}) by substituting $\mathbf{\Theta}\mathbf{X}$ for $\mathbf{Z}$
\begin{equation}
\label{eq3}
\begin{aligned}
       \mathop{\min}_{\mathbf{\Theta}}\tr(\mathbf{\Theta}\mathbf{X}\mathbf{L}\mathbf{X}^{\T}\mathbf{\Theta}^{\T}), \quad \mathrm{s.t.} \quad \mathbf{\Theta}\mathbf{X}\mathbf{D}\mathbf{X}^{\T}\mathbf{\Theta}^{\T}=\mathbf{I},
\end{aligned}
\end{equation}
which can be solved by generalized eigenvalue decomposition.

Different from the previously mentioned subspace learning methods, a re-gression-based joint learning model \cite{ji2009linear} can explicitly bridge the learned latent subspace and labels, which can be formulated in a general form:
\begin{equation}
\label{eq4}
\begin{aligned}
       \mathop{\min}_{\mathbf{P},\mathbf{\Theta}}\frac{1}{2}\mathbf{E}(\mathbf{P},\mathbf{\Theta})+\frac{\beta}{2}\mathbf{\Phi}(\mathbf{\Theta})+\frac{\gamma}{2}\mathbf{\Psi}(\mathbf{P}),
\end{aligned}
\end{equation}
where $\mathbf{E}(\mathbf{P},\mathbf{\Theta})$ is the error term defined as $\norm{\mathbf{Y}-\mathbf{P}\mathbf{\Theta}\mathbf{X}}_{\F}^{2}$, $\norm{\bullet}_{\F}$ represents a Frobenius norm, $\beta$ and $\gamma$ are the corresponding penalty parameters. $\mathbf{\Phi}$ and $\mathbf{\Psi}$ denote regularization functions, which might be $l_{1}$ norm, $l_{2}$ norm, $l_{2,1}$ norm or manifold regularization. Herein, the variable $\mathbf{\Theta}$ is called intermediate transformation and the corresponding subspace generated by $\mathbf{\Theta}$ is called latent subspace where the feature can be further structurally learned and represented in a more suitable way \cite{Hu2016}.

On the basis of Eq. (\ref{eq5}), we further extend the framework by following a progressive learning strategy:
\begin{equation}
\label{eq5}
\begin{aligned}
       \mathop{\min}_{\mathbf{P},\{\mathbf{\Theta}_{l}\}_{l=1}^{m}}\frac{1}{2}\mathbf{E}(\mathbf{P},\{\mathbf{\Theta}_{l}\}_{l=1}^{m})+\frac{\beta}{2}\mathbf{\Phi}(\{\mathbf{\Theta}_{l}\}_{l=1}^{m})+\frac{\gamma}{2}\mathbf{\Psi}(\mathbf{P}),
\end{aligned}
\end{equation}
where $\mathbf{E}(\mathbf{P},\{\mathbf{\Theta}_{l}\}_{l=1}^{m})$ is specified as $\norm{\mathbf{Y}-\mathbf{P}\mathbf{\Theta}_{m}...\mathbf{\Theta}_{l}...\mathbf{\Theta}_{1}\mathbf{X}}_{\F}^{2}$ and $\{\mathbf{\Theta}_{l}\}_{l=1}^{m}$ represent a set of intermediate transformations.
\begin{figure*}[!t]
\label{Fig2}
\centering
        \includegraphics[width=12cm]{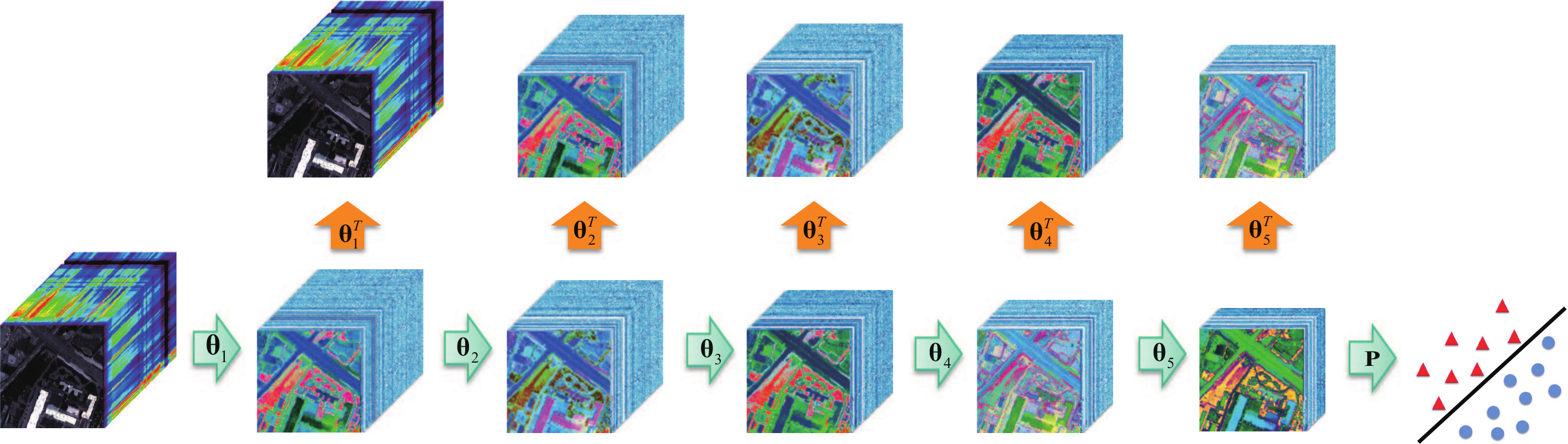}
\caption{The illustration of the proposed J-Play framework.}
\end{figure*}
\subsection{Problem Formulation}
Following the general framework given in Eq.(\ref{eq6}), the proposed J-Play can be formulated as the following constrained optimization problem:
\begin{equation}
\label{eq6}
\begin{aligned}
       \mathop{\min}_{\mathbf{P},\{\mathbf{\Theta}_{l}\}_{l=1}^{m}}&\frac{1}{2}\mathbf{\Upsilon}(\{\mathbf{\Theta}_{l}\}_{l=1}^{m})
       +\frac{\alpha}{2}\mathbf{E}(\mathbf{P},\{\mathbf{\Theta}_{l}\}_{l=1}^{m})+\frac{\beta}{2}\mathbf{\Phi}(\{\mathbf{\Theta}_{l}\}_{l=1}^{m})+\frac{\gamma}{2}\mathbf{\Psi}(\mathbf{P})\\
       &\mathrm{s.t.} \quad  \mathbf{X}_{l}=\mathbf{\Theta}_{l}\mathbf{X}_{l-1}, \quad \mathbf{X}_{l}\succeq 0, \quad \norm{\mathbf{x}_{lk}}_{2} \preceq 1, \quad \forall l=1,2,...,m,
\end{aligned}
\end{equation}
where $\mathbf{X}$ is assigned to $\mathbf{X}_{0}$, while $\alpha$, $\beta$, and $\gamma$  are three penalty parameters corresponding to the different terms, which aim at balancing the importance between the terms. Fig. 2 illustrates the J-Play framework. Since Eq. (\ref{eq7}) is a typically ill-posed problem, reasonable assumptions or priors need to be introduced to search a solution in a narrowed range effectively. More specifically, we cast Eq.(\ref{eq7}) as a least-square regression problem with reconstruction loss term ($\mathbf{\Upsilon}(\bullet)$), prediction loss term ($\mathbf{E}(\bullet)$) and two regularization terms ($\mathbf{\Phi}(\bullet)$ and $\mathbf{\Psi}(\bullet)$). We detail these terms one by one as follows.

\emph{1) Reconstruction Loss Term }$\mathbf{\Upsilon}(\{\mathbf{\Theta}_{l}\}_{l=1}^{m})$: Without any constraints or prior, directly estimating multi-coupled projections in J-Play is hardly performed with the increase of the number of estimated projections. This can be reasonably explained by gradient missing between the two neighboring variables estimated in the process of optimization. That is, the variations between these neighboring projections are made to be tiny and even zero. In particular, when the number of projections increases to a certain extent, most of learned projections tend to be zero and become meaningless. To this end, we adopt a kind of autoencoder-like scheme to make the learned subspace projected back to the original space as much as possible. The benefits of the scheme are, on one hand, to prevent the data over-fitting to some extent, especially avoiding overmuch noises from being considered; on the other hand, to establish an effective link between the original space and the subspace, making the learned subspace more meaningful. Therefore, the resulting expression is
\begin{equation}
\label{eq7}
\begin{aligned}
      \mathbf{\Upsilon}(\{\mathbf{\Theta}_{l}\}_{l=1}^{m})=\sum\nolimits_{l=1}^{m}\norm{\mathbf{X}_{l-1}-\mathbf{\Theta}_{l}^{T}\mathbf{\Theta}_{l}\mathbf{X}_{l-1}}_{\F}^{2}.
\end{aligned}
\end{equation}
In our case, to fully utilize the advantages of this term, we consider it in each latent subspace as shown in Eq.(\ref{eq8}).

\emph{2) Predication Loss Term }$\mathbf{E}(\mathbf{P},\{\mathbf{\Theta}_{l}\}_{l=1}^{m})$: This term is to minimize the empirical risk between the original data and the corresponding labels through multi-coupled projections in a progressive way, which can be formulated as
\begin{equation}
\label{eq8}
\begin{aligned}
      \mathbf{E}(\mathbf{P},\{\mathbf{\Theta}_{l}\}_{l=1}^{m})=\norm{\mathbf{Y}-\mathbf{P}\mathbf{\Theta}_{m}...\mathbf{\Theta}_{l}...\mathbf{\Theta}_{1}\mathbf{X}}_{\F}^{2}.
\end{aligned}
\end{equation}

\emph{3) Local Manifold Regularization }$\mathbf{\Phi}(\{\mathbf{\Theta}_{l}\}_{l=1}^{m})$: As introduced in \cite{Wang2016}, a manifold structure is an important prior for subspace learning. Superior to vector-based feature learning, such as artificial neural network (ANN), a manifold structure can effectively capture the intrinsic structure between samples. To facilitate structure learning in J-Play, we perform the local manifold regularization to each latent subspace. Specifically, this term can be expressed by
\begin{equation}
\label{eq9}
\begin{aligned}
      \mathbf{\Phi}(\{\mathbf{\Theta}_{l}\}_{l=1}^{m})=\sum\nolimits_{l=1}^{m}\tr(\mathbf{\Theta}_{l}\mathbf{X}_{l-1}\mathbf{L}\mathbf{X}_{l-1}^{\T}\mathbf{\Theta}_{l}^{\T}).
\end{aligned}
\end{equation}
\begin{algorithm}[!t]
\scriptsize
\caption{Joint \& Progressive Learning Strategy (J-Play)}
\KwIn{$\mathbf{Y}, \mathbf{X},\mathbf{L},$ and parameters $\alpha, \beta, \gamma$ and $maxIter.$}
\KwOut{$\{\mathbf{\Theta}_{l}\}_{l=1}^{m}.$}
{$\textbf{Initialization Step: }$}\\
{Greedily initialize $\mathbf{\Theta}_{l}$ corresponding to each latent subspace}:\\
\For{$l=1:m$}
{
  $\mathbf{\Theta}^{0}_{l} \leftarrow LPP(\mathbf{X}_{l-1})$\\
  $\mathbf{\Theta}_{l} \leftarrow AutoRULe(\mathbf{X}_{l-1},\mathbf{\Theta}^{0}_{l},\mathbf{L})$\\
  $\mathbf{X}_{l} \leftarrow \mathbf{\Theta}_{l}\mathbf{X}_{l-1}$\\
}
{$\textbf{Fine-tuning Step: }$}\\
  $t=0,\zeta=1e-4;$\\
  \While{not converged \rm{or} $t>maxIter$}
 {
   Fix other variables to update $\mathbf{P}$ by solving a subproblem of $\mathbf{P}$;\\
   \For{$i=1:m$}
   {
      Fix other variables to update $\mathbf{\Theta}_{l}^{t+1}$ by solving a subproblem of $\mathbf{\Theta}_{l};$
   }
   Compute the objective function value $Obj^{t+1}$ and check the convergence condition:
   \eIf{$|\frac{Obj^{t+1}-Obj^{t}}{Obj^{t}}|<\zeta$}
   {
     Stop iteration;
   }
   {
     $t\leftarrow t+1$;
   }
 }
\end{algorithm}
\emph{4) Regression Coefficient Regularization }$\mathbf{\Psi}(\mathbf{P})$: The regularization term can promote us to derive a more reasonable solution with a reliable generalization to our model, which can be written as
\begin{equation}
\label{eq10}
\begin{aligned}
      \mathbf{\Psi}(\mathbf{P})=\norm{\mathbf{P}}_{\F}^{2}.
\end{aligned}
\end{equation}

Moreover, the non-negativity constraint with respect to each learned dimension-reduced feature (e.g. $\{\mathbf{X}_{l}\}_{l=1}^{m} \succeq 0$) is considered since we aim to obtain a meaningful low-dimensional feature representation similar to original image data acquired in a non-negative unit. In addition to the non-negativity constraint, we also impose a norm constraint \footnote{Regarding this constraint,please refer to \cite{Lee2007} for more details.} for sample-based of each subspace: $\norm{\mathbf{x}_{lk}}_{2} \preceq 1, \forall k=1,...,N$ and $l=1,...,m$.
\subsection{Model Optimization}
Considering the complexity and the non-convexity of our model, we pretrain our model to have an initial approximation of subspace projections $\{\mathbf{\Theta}_{l}\}_{l=1}^{m}$ as this can greatly reduce the model's training time and also help finding an optimal solution easier. This is a common tactic that has been successfully employed in deep autoencoders \cite{Hinton2006}. Inspired by this trick, we propose a pre-training model with respect to $\mathbf{\Theta}_{l},\forall l=1,...,m$ by simplifying Eq.(\ref{eq7}) as
\begin{equation}
\label{eq11}
\begin{aligned}
       \mathop{\min}_{\mathbf{\Theta}_{l}}\frac{1}{2}\mathbf{\Upsilon}(\mathbf{\Theta}_{l})+\frac{\eta}{2}\mathbf{\Phi}(\mathbf{\Theta}_{l})
       \quad \mathrm{s.t.} \quad \mathbf{X}_{l}\succeq 0, \quad \norm{\mathbf{x}_{lk}}_{2} \preceq 1,
\end{aligned}
\end{equation}
which is named as \textbf{auto}-\textbf{r}econstructing \textbf{u}nsupervised \textbf{le}arning (AutoRULe). Given the outputs of AutoRULe, the problem of Eq. (\ref{eq7}) can be more effectively solved by an alternatively minimizing strategy that separately solves two subproblems with respect to $\{\mathbf{\Theta}_{l}\}_{l=1}^{m}$ and $\mathbf{P}$. Therefore, the global algorithm of J-Play can be summarized in \textbf{Algorithm 1},where AutoRULe is initialized by LPP.

The pre-training method (AutoRULe) can be effectively solved via the ADMM-based framework. Following this, we consider an equivalent form of Eq. (\ref{eq12}) by introducing multiple auxiliary variables $\mathbf{H}$, $\mathbf{G}$, $\mathbf{Q}$ and $\mathbf{S}$ to replace $\mathbf{X}_{l}$, $\mathbf{\Theta}_{l}$, $\mathbf{X}_{l}^{+}$ and $\mathbf{X}_{l}^{\sim}$, respectively, where $()^{+}$ denotes an operator that converts each component of the matrix to its absolute value and $()^{\sim}$ is a proximal operator for solving the constraint of $\norm{\mathbf{x}_{lk}}_{2} \preceq 1$ \cite{Heide2015}, written as follows
\begin{equation}
\label{eq12}
\begin{aligned}
       \mathop{\min}_{\mathbf{\Theta}_{l},\mathbf{H},\mathbf{G},\mathbf{Q},\mathbf{S}}&\frac{1}{2}\mathbf{\Upsilon}(\mathbf{G},\mathbf{H})+\frac{\eta}{2}\mathbf{\Phi}(\mathbf{\Theta}_{l})
       =\frac{1}{2}\norm{\mathbf{X}_{l-1}-\mathbf{G}^{\T}\mathbf{H}}_{\F}^{2}+\frac{\eta}{2}\tr(\mathbf{X}_{l}\mathbf{L}\mathbf{X}_{l}^{\T})\\
       &\mathrm{s.t.} \quad \mathbf{Q}\succeq 0, \quad \norm{\mathbf{s}_{k}}_{2} \preceq 1, \quad \mathbf{X}_{l}=\mathbf{\Theta}_{l}\mathbf{X}_{l-1},\\
       &\quad \quad \quad \mathbf{X}_{l}=\mathbf{H}, \quad \mathbf{\Theta}_{l}=\mathbf{G}, \quad \mathbf{X}_{l}=\mathbf{Q}, \quad \mathbf{X}_{l}=\mathbf{S}.
\end{aligned}
\end{equation}
The augmented Lagrangian version of Eq. (\ref{eq13}) is
\begin{equation}
\label{eq13}
\begin{aligned}
      \mathscr{L}&_{\mu}\left(\mathbf{\Theta}_{l},\mathbf{H},\mathbf{G},\mathbf{Q},\mathbf{S}, \{\mathbf{\Lambda}_{n}\}_{n=1}^{4} \right)\\ &=\frac{1}{2}\norm{\mathbf{X}_{l-1}-\mathbf{G}^{\T}\mathbf{H}}_{\F}^{2}+\frac{\eta}{2}\tr(\mathbf{\Theta}_{l}\mathbf{X}_{l-1}\mathbf{L}\mathbf{X}_{l-1}^{\T}\mathbf{\Theta}_{l}^{\T})
      +\mathbf{\Lambda}_{1}^{\T}(\mathbf{H}-\mathbf{\Theta}_{l}\mathbf{X}_{l-1})\\
      &+\mathbf{\Lambda}_{2}^{\T}(\mathbf{G}-\mathbf{\Theta}_{l})+\mathbf{\Lambda}_{3}^{\T}(\mathbf{Q}-\mathbf{\Theta}_{l}\mathbf{X}_{l-1})+\mathbf{\Lambda}_{4}^{\T}(\mathbf{S}-\mathbf{\Theta}_{l}\mathbf{X}_{l-1})+\frac{\mu}{2}\norm{\mathbf{H}-\mathbf{\Theta}_{l}\mathbf{X}_{l-1}}_{\F}^{2}\\
      &+\frac{\mu}{2}\norm{\mathbf{G}-\mathbf{\Theta}_{l}}_{\F}^{2}+\frac{\mu}{2}\norm{\mathbf{Q}-\mathbf{\Theta}_{l}\mathbf{X}_{l-1}}_{\F}^{2}+\frac{\mu}{2}\norm{\mathbf{S}-\mathbf{\Theta}_{l}\mathbf{X}_{l-1}}_{\F}^{2}+l_{R}^{+}(\mathbf{Q})+l_{R}^{\sim}(\mathbf{S}),
\end{aligned}
\end{equation}
where $\{\mathbf{\Lambda}_{n}\}_{n=1}^{4}$ are Lagrange multipliers and $\mu$ is the penalty parameter. The two terms $l_{R}^{+}(\bullet)$ and $l_{R}^{\sim}(\bullet)$ represent two kinds of projection operators, respectively. That is, $l_{R}^{+}(\bullet)$ is defined as
\begin{equation}
\label{eq14}
   max(\bullet)=
   \begin{aligned}
       \begin{cases}
         \quad \bullet \;, \quad \bullet \succ 0\\
         \quad 0 \;, \quad \bullet \preceq 0,   
       \end{cases}
   \end{aligned}
\end{equation}
while $l_{R}^{\sim}(\bullet_{k})$ is a vector-based operator defined by
\begin{equation}
\label{eq15}
\begin{aligned}
      prox_{f}(\bullet_{k})=
       \begin{cases}
       \; \frac{\bullet_{k}}{\norm{\bullet_{k}}_{2}} \;, \quad \norm{\bullet_{k}}_{2} \succ 1\\
       \quad \bullet_{k} \;, \quad \norm{\bullet_{k}}_{2} \preceq 1,
       \end{cases}
\end{aligned}
\end{equation}
where $\bullet_{k}$ is the $k$th column of matrix $\bullet$. \textbf{Algorithm 2} details the procedures of AutoRULe.
\begin{algorithm}[!t]
\scriptsize
\caption{Auto-reconstructing unsupervised learning (AutoRULe)}
\KwIn{$\mathbf{X}_{l-1},\mathbf{\Theta}^{0}_{l},\mathbf{L},$ and parameters $\eta$ and $maxIter$.}
\KwOut{$\mathbf{\Theta}_{l}.$}
\textbf{Initialization}: $\mathbf{H}^{0}=\mathbf{\Theta}^{0}_{l}\mathbf{X}_{l-1}, \mathbf{G}^{0}=\mathbf{0}, \mathbf{Q}^{0}=\mathbf{P}^{0}=\mathbf{0}, \mathbf{\Lambda}_{2}^{0}=\mathbf{0}, \mathbf{\Lambda}_{1}^{0}=\mathbf{\Lambda}_{3}^{0}=\mathbf{\Lambda}_{4}^{0}=\mathbf{0}, \mu^{0}=1e-3, \mu_{max}=1e6, \rho=2, \varepsilon=1e-6, t=0.$\\
  \While{not converged \rm{or} $t>maxIter$}
 {
         Fix $\mathbf{H}^{t}, \mathbf{G}^{t}, \mathbf{Q}^{t}, \mathbf{P}^{t}$ to update $\mathbf{\Theta}^{t+1}_{l}$ by
         {\setlength\abovedisplayskip{1pt}
         \setlength\belowdisplayskip{1pt}
         \begin{equation*}
         \begin{aligned}
                \mathbf{\Theta}_{l}=&(\mu\mathbf{H}\mathbf{X}_{l-1}^{T}+\mathbf{\Lambda}_{1}\mathbf{X}_{l-1}^{T}+\mu\mathbf{G}+\mathbf{\Lambda}_{2}+\mu\mathbf{Q}\mathbf{X}_{l-1}^{T}+\mathbf{\Lambda}_{3}\mathbf{X}_{l-1}^{T}\\
                &+\mu\mathbf{P}\mathbf{X}_{l-1}^{T}+\mathbf{\Lambda}_{4}\mathbf{X}_{l-1}^{T})(\eta(\mathbf{X}_{l-1}\mathbf{L}\mathbf{X}_{l-1}^{T})+3\mu(\mathbf{X}_{l-1}\mathbf{X}_{l-1}^{T})+\mu\mathbf{I})^{-1}.
         \end{aligned}
         \end{equation*}}\\
         Fix $\mathbf{\Theta}^{t+1}_{l}, \mathbf{G}^{t}, \mathbf{Q}^{t}, \mathbf{P}^{t}$ to update $\mathbf{H}^{t+1}$ by
         {\setlength\abovedisplayskip{1pt}
         \setlength\belowdisplayskip{1pt}
         \begin{equation*}
         \begin{aligned}
                \mathbf{H}=(\mathbf{G}\mathbf{G}^{T}+\mu\mathbf{I})^{-1}(\mathbf{G}\mathbf{X}_{l-1}+\mu\mathbf{\Theta}_{l}\mathbf{X}_{l-1}-\mathbf{\Lambda}_{1}).
         \end{aligned}
         \end{equation*}}\\
         Fix $\mathbf{H}^{t+1}, \mathbf{\Theta}^{t+1}_{l}, \mathbf{Q}^{t}, \mathbf{P}^{t}$ to update $\mathbf{G}^{t+1}$ by
         {\setlength\abovedisplayskip{1pt}
         \setlength\belowdisplayskip{1pt}
         \begin{equation*}
         \begin{aligned}
                \mathbf{G}=(\mathbf{H}\mathbf{H}^{T}+\mu\mathbf{I})^{-1}(\mathbf{H}\mathbf{X}_{i}+\mu\mathbf{\Theta}_{l}-\mathbf{\Lambda}_{2}).
         \end{aligned}
         \end{equation*}}\\
         Fix $\mathbf{H}^{t+1}, \mathbf{G}^{t+1}, \mathbf{\Theta}^{t+1}_{l}, \mathbf{P}^{t}$ to update $\mathbf{Q}^{t+1}$ by
         {\setlength\abovedisplayskip{1pt}
         \setlength\belowdisplayskip{1pt}
         \begin{equation*}
         \begin{aligned}
                \mathbf{Q}=max(\mathbf{\Theta}_{l}\mathbf{X}_{l-1}-\mathbf{\Lambda}_{3}/\mu,0).
         \end{aligned}
         \end{equation*}}\\
         Fix $\mathbf{H}^{t+1}, \mathbf{G}^{t+1}, \mathbf{\Theta}^{t+1}_{l}, \mathbf{Q}^{t+1}$ to update $\mathbf{P}^{t+1}$ by
         {\setlength\abovedisplayskip{1pt}
         \setlength\belowdisplayskip{1pt}
         \begin{equation*}
         \begin{aligned}
                \mathbf{P}=prox_{f}(\mathbf{\Theta}_{l}\mathbf{X}_{l-1}-\mathbf{\Lambda}_{4}/\mu).
         \end{aligned}
         \end{equation*}}\\
         Update Lagrange multipliers by
         {\setlength\abovedisplayskip{1pt}
         \setlength\belowdisplayskip{1pt}
         \begin{equation*}
         \begin{aligned}
                \mathbf{\Lambda}_{1}^{t+1}&=\mathbf{\Lambda}_{1}^{t}+\mu^{t}(\mathbf{H}^{t+1}-\mathbf{\Theta}_{i}^{t+1}\mathbf{X}_{l-1}), \mathbf{\Lambda}_{2}^{t+1}=\mathbf{\Lambda}_{2}^{t}+\mu^{t}(\mathbf{G}^{t+1}-\mathbf{\Theta}_{i}^{t+1}),\\
                \mathbf{\Lambda}_{3}^{t+1}&=\mathbf{\Lambda}_{3}^{t}+\mu^{t}(\mathbf{Q}^{t+1}-\mathbf{\Theta}_{i}^{t+1}\mathbf{X}_{l-1}),
                \mathbf{\Lambda}_{4}^{t+1}=\mathbf{\Lambda}_{4}^{t}+\mu^{t}(\mathbf{P}^{t+1}-\mathbf{\Theta}_{i}^{t+1}\mathbf{X}_{l-1}).
         \end{aligned}
         \end{equation*}}\\
         Update penalty parameter by
         {\setlength\abovedisplayskip{1pt}
         \setlength\belowdisplayskip{1pt}
         \begin{equation*}
         \begin{aligned}
         \mu^{t+1}=min(\rho\mu^{t},\mu_{max}).
         \end{aligned}
         \end{equation*}}\\
         Check the convergence conditions:
         \eIf{$\norm {\mathbf{H}^{t+1}-\mathbf{\Theta}_{l}^{t+1}\mathbf{X}_{l-1}}_{F}<\varepsilon$ and $\norm {\mathbf{G}^{t+1}-\mathbf{\Theta}_{l}^{t+1}}_{F}<\varepsilon$ and $\norm {\mathbf{Q}^{t+1}-\mathbf{\Theta}_{l}^{t+1}\mathbf{X}_{l-1}}_{F}<\varepsilon$ and $\norm {\mathbf{P}^{t+1}-\mathbf{\Theta}_{l}^{t+1}\mathbf{X}_{l-1}}_{F}<\varepsilon$}
         {
           Stop iteration;
         }
         {
         $t\leftarrow t+1$;
         }
 }
\end{algorithm}

The two subproblems in \textbf{Algorithm 1} can be optimized alternatively as follows:

\emph{Optimization with respect to $\mathbf{P}$}: This is a typical least square regression problem, which can be written as
\begin{equation}
\label{eq16}
\begin{aligned}
       \mathop{\min}_{\mathbf{P}}\frac{\alpha}{2}\mathbf{E}(\mathbf{P})+\frac{\gamma}{2}\mathbf{\Psi}(\mathbf{P})=\frac{\alpha}{2}\norm{\mathbf{Y}-\mathbf{P}\mathbf{\Theta}_{m}...\mathbf{\Theta}_{l}...\mathbf{\Theta}_{1}\mathbf{X}}_{\F}^{2}
       +\frac{\gamma}{2}\norm{\mathbf{P}}_{\F}^{2},
\end{aligned}
\end{equation}
which has a closed-form solution
\begin{equation}
\label{eq17}
\begin{aligned}
       \mathbf{P} \leftarrow (\alpha\mathbf{Y}\mathbf{V}^{\T})(\alpha\mathbf{V}\mathbf{V}^{\T}+\gamma\mathbf{I})^{-1},
\end{aligned}
\end{equation}
where $\mathbf{V}=\mathbf{\Theta}_{m}...\mathbf{\Theta}_{l}...\mathbf{\Theta}_{1}, \forall l=1,...,m$.

\emph{Optimization with respect to $\{\mathbf{\Theta}_{l}\}_{l=1}^{m}$}: The variables $\{\mathbf{\Theta}_{l}\}_{l=1}^{m}$ can be individually optimized, and hence the optimization problem of each $\mathbf{\Theta}_{l}$ can be generally formulated by
\begin{equation}
\label{eq18}
\begin{aligned}
       \mathop{\min}_{\mathbf{\Theta}_{l}}&\frac{1}{2}\mathbf{\Upsilon}(\mathbf{\Theta}_{l})+\frac{\alpha}{2}\mathbf{E}(\mathbf{\Theta}_{l})+\frac{\beta}{2}\mathbf{\Phi}(\mathbf{\Theta}_{l})
       =\frac{1}{2}\norm{\mathbf{X}_{l-1}-\mathbf{\Theta}_{l}^{\T}\mathbf{\Theta}_{l}\mathbf{X}_{l-1}}_{\F}^{2}\\
       &+\frac{\alpha}{2}\norm{\mathbf{Y}-\mathbf{P}\mathbf{\Theta}_{m}...\mathbf{\Theta}_{l}...\mathbf{\Theta}_{1}\mathbf{X}}_{\F}^{2}
       +\frac{\beta}{2}\tr(\mathbf{\Theta}_{l}\mathbf{X}_{l-1}\mathbf{L}\mathbf{X}_{l-1}^{\T}\mathbf{\Theta}_{l}^{\T})\\
       &\mathrm{s.t.} \quad \mathbf{X}_{l}=\mathbf{\Theta}_{l}\mathbf{X}_{l-1}, \quad \mathbf{X}_{l}\succeq 0, \quad \norm{\mathbf{x}_{lk}}_{2} \preceq 1,
\end{aligned}
\end{equation}
which can be basically deduced by following the framework of \textbf{Algorithm 2}. The only difference lies in the optimization subproblem with respect to $\mathbf{H}$ whose solution can be collected by solving the following problem:
\begin{equation}
\label{eq19}
\begin{aligned}
       \mathop{\min}_{\mathbf{H}}&\frac{1}{2}\norm{\mathbf{X}_{l-1}-\mathbf{G}^{\T}\mathbf{H}}_{\F}^{2}+\frac{\alpha}{2}\norm{\mathbf{Y}-\mathbf{P}_{l}\mathbf{H}}_{\F}^{2}
       +\mathbf{\Lambda}_{1}^{\T}(\mathbf{H}-\mathbf{\Theta}_{l}\mathbf{X}_{l-1})\\
       &+\frac{\mu}{2}\norm{\mathbf{H}-\mathbf{\Theta}_{l}\mathbf{X}_{l-1}}_{\F}^{2} \quad \mathrm{s.t.} \quad \mathbf{P}_{l}=\mathbf{P}_{l-1}\mathbf{\Theta}_{l+1}, \quad \mathbf{P}_{0}=\mathbf{P}.
\end{aligned}
\end{equation}
The analytical solution of Eq. (\ref{eq20}) is given by
\begin{equation}
\label{eq20}
\begin{aligned}
       \mathbf{H} \leftarrow (\alpha\mathbf{P}_{l}^{\T}\mathbf{P}_{l}+\mathbf{G}\mathbf{G}^{\T}+\mu\mathbf{I})^{-1}
       (\alpha\mathbf{P}_{l}^{\T}\mathbf{Y}+\mathbf{G}\mathbf{X}_{l-1}+\mu\mathbf{\Theta}_{l}\mathbf{X}_{l-1}-\mathbf{\Lambda}_{1}).
\end{aligned}
\end{equation}
Finally, we repeat these optimization procedures until a stopping criterion is satisfied. Please refer to \textbf{Algorithm 1} and \textbf{Algorithm 2} for more explicit steps.

\section{Experiments}
In this section, we conduct the classification to quantitatively evaluate the performance of the proposed method (J-Play) using three popular and advanced classifiers, namely the nearest neighbor (NN) based on the Euclidean distance, kernel support vector machines (KSVM) and canonical correlation forest (CCF), in comparison with previous state-of-the-art methods. Overall accuracy (OA) is given to quantify the classification performance.
\subsection{Data Description}
The experiments are performed on two different types of datasets: hyperspectral datasets and face datasets, as both of them easily suffer from the information redundancy and need to improve the representative ability of features. We have used the following two hyperspectral datasets and two face datasets:

1) \emph{Indian Pines AVIRIS Image:} The first hyperspectral cube was acquired by the AVIRIS sensor with the size of $145\times145\times220$, which consists of $16$ class of vegetation. More specific classes and the arrangement of training and test samples can be found in \cite{hong2016k}. The first image of Fig. 3 shows a false color image of Indian Pines data.

2) \emph{University of Houston Image:} The second hyperspectral cube was provided for the $2013$ IEEE GRSS data fusion contest acquired by ITRES-CASI sensor with size of $349\times1905\times144$. The information regarding classes and corresponding train and test samples can be found in \cite{Hong2017}. A false color image of the study scene is shown in the first image of Fig. 4.

3) \emph{Extended Yale-B Dataset:} We only choose a subset of the mentioned dataset with the frontal pose and the different illuminations of $38$ subjects ($2414$ images in total), which can widely used in evaluating the performance of subspace learning \cite{ZhangL2011}\cite{Cai2007}. These images were aligned and cropped to the size of $32\times32$, that is, $1024$-dimensional vector-based representation. Each individual has $64$ near frontal images under different illuminations.

4) \emph{ AR Dataset:} Similar to \cite{Yang2011}, we choose a subset of AR under the conditions of illumination and expressions, which comprises of $100$ subjects. Each person has $14$ images with seven ones from Session $1$ as training set and others from Session $2$ as testing samples. The images are resized to $60\times43$.
\subsection{Experimental Steup}
As the fixed training and testing samples are given for the hyperspectral datasets, subspace learning techniques can directly be performed on training set to learn an optimal subspace where the testing set can be simply classified by NN, KSVM, and CCF. For the face datasets, since there is no standard training and testing sets, ten replications are performed for randomly selecting training and testing samples. A random subset with $10$ facial images per individual is chosen with labels as the training set and the rest of it is considered to be the testing set. Furthermore, we compare the performance of the proposed method (J-Play) with the baseline (original features without dimensionality reduction) and six popular and advanced methods (PCA, LPP, LDA, LFDA, LSDR, and LSQMID). With learning the different number of coupled projections, the proposed method can be successively specified as J-Play$_{1}$,...,J-Play$_{l}$,...,J-Play$_{m}$, $\forall l=1,...,m$. To investigate the trend of OAs, $m$ are uniformly set up to $7$ on the four datasets.
\begin{figure*}[!t]
	  \centering
		{
	     \includegraphics[width=12cm]{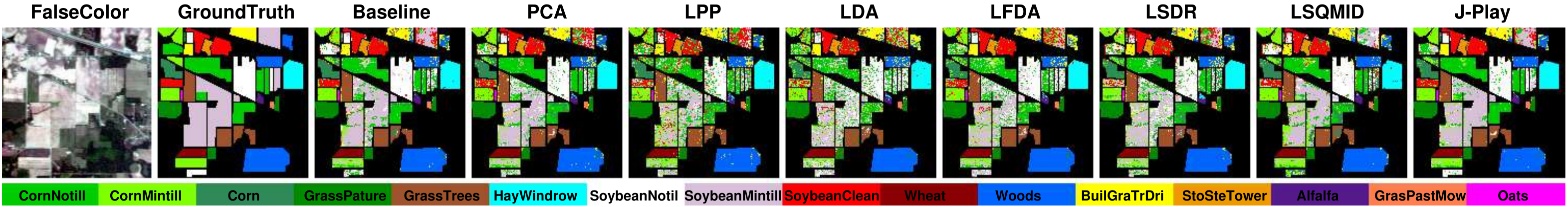}
		}	
         \caption{A false color image, ground truth and classification maps of the different algorithms obtained using CCF on the Indian Pines dataset.}
\end{figure*}
\begin{figure*}[!t]
	  \centering
		{
	     \includegraphics[width=12cm]{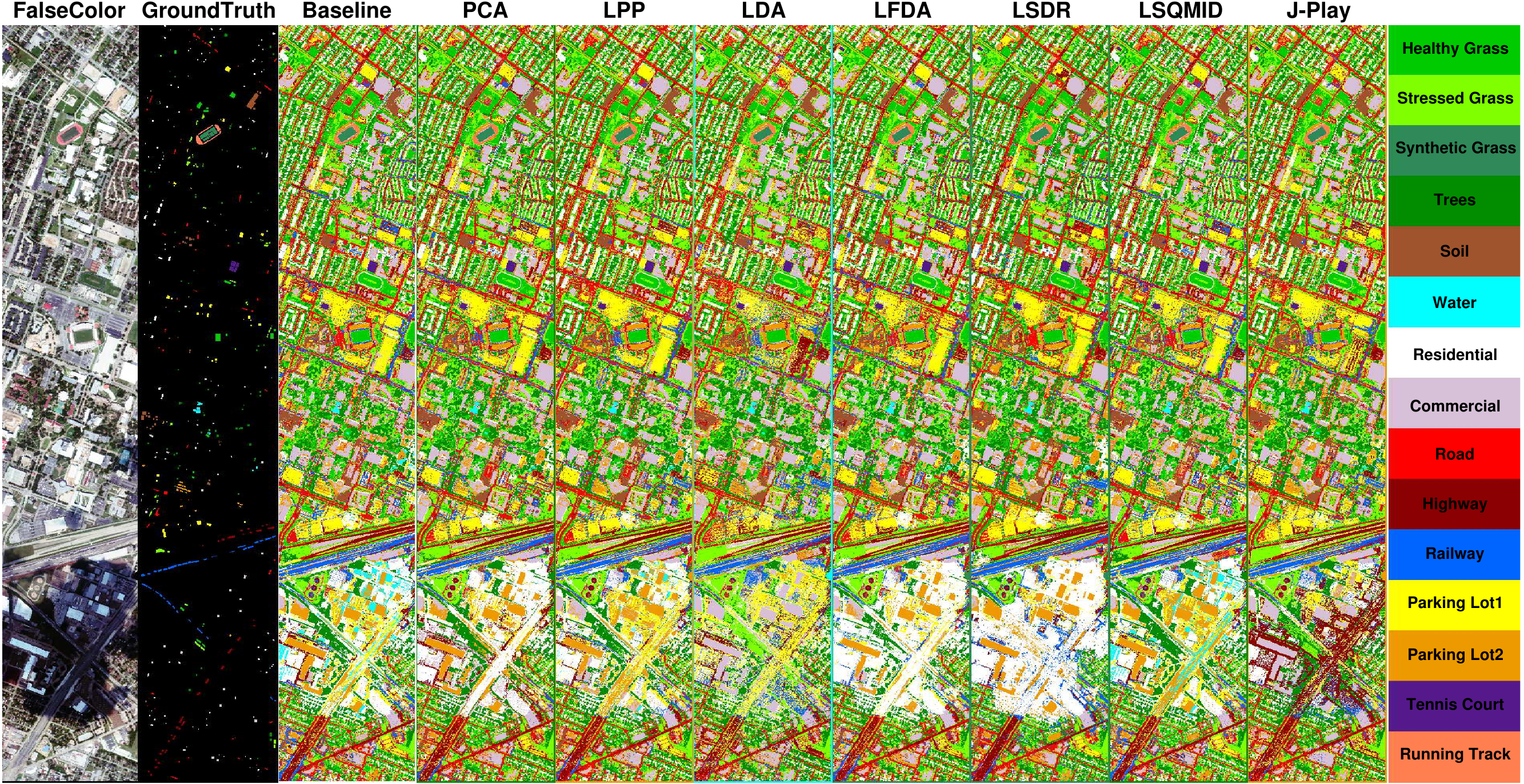}
		}	
         \caption{A false color image, ground truth and classification maps of the different algorithms obtained using CCF on the Houston dataset.}
\end{figure*}
\begin{table}[!t]
\begin{center}
\caption{Quantitative performance comparisons on two hyperspectral datasets. The best results for the different classifiers are shown in red.}
\label{table:headings}
\begin{tabular}{ccccccc}
\hline\noalign{\smallskip}
\multirow{2}{*}{\bf Methods} & \multicolumn{3}{c}{Indian Pines dataset} & \multicolumn{3}{c}{Houston dataset} \\
\cline{2-7} & NN & KSVM & CCF & NN & KSVM & CCF\\
\hline\noalign{\smallskip}
 Baseline $(220/144)$ & $65.89\%$ & $66.56\%$ & $81.71\%$ & $72.83\%$ & $80.19\%$ & $82.60\%$\\
\hline\noalign{\smallskip}
 PCA $(20/20)$ & $65.40\%$ & $75.25\%$ & $79.26\%$ & $72.75\%$ & $79.54\%$ & $83.90\%$\\
\hline\noalign{\smallskip}
 LPP $(20/30)$ & $64.86\%$ & $63.02\%$ & $68.48\%$ & $75.31\%$ & $78.43\%$ & $81.77\%$\\
\hline\noalign{\smallskip}
 LDA $(15/14)$ & $64.14\%$ & $63.88\%$ & $65.61\%$ & $75.81\%$ & $76.66\%$ & $79.62\%$\\
\hline\noalign{\smallskip}
 LFDA $(15/14)$ & $73.86\%$ & $74.25$\% & $75.17\%$ & $75.52\%$ & $80.46\%$ & $82.27\%$\\
\hline\noalign{\smallskip}
 LSDR $(50/40)$ & $73.67\%$ & $76.84\%$ & $77.38\%$ & $76.80\%$ & $80.39\%$ & $81.64\%$\\
\hline\noalign{\smallskip}
 LSQMID $(60/80)$ & $66.94\%$ & $78.90\%$ & $79.32\%$ & $76.31\%$ & $80.23\%$ & $81.69\%$\\
\hline\noalign{\smallskip}
 J-Play$_{1}$ $(20/30)$ & $78.81\%$ & $82.04\%$ & $82.24\%$ & $78.22\%$ & $83.32\%$ & $85.09\%$\\
\hline\noalign{\smallskip}
 J-Play$_{2}$ $(20/30)$ & $80.87\%$ & $83.75\%$ & $83.23\%$ & $79.16\%$ & $\color{red}84.41\%$ & $85.15\%$\\
\hline\noalign{\smallskip}
 J-Play$_{3}$ $(20/30)$ & $83.59\%$ & $85.08\%$ & $84.44\%$ & $\color{red}80.13\%$ & $83.68\%$ & $\color{red}88.19\%$\\
\hline\noalign{\smallskip}
 J-Play$_{4}$ $(20/30)$ & $\color{red}83.92\%$ & $85.21\%$ & $\color{red}84.57\%$ & $79.64\%$ & $83.25\%$ & $85.63\%$\\
\hline\noalign{\smallskip}
 J-Play$_{5}$ $(20/30)$ & $83.76\%$ & $\color{red}85.30\%$ & $84.41\%$ & $80.00\%$ & $82.21\%$ & $85.81\%$\\
\hline\noalign{\smallskip}
 J-Play$_{6}$ $(20/30)$ & $83.56\%$ & $84.79\%$ & $83.82\%$ & $79.69\%$ & $82.45\%$ & $84.82\%$\\
\hline\noalign{\smallskip}
 J-Play$_{7}$ $(20/30)$ & $82.70\%$ & $83.82\%$ & $83.04\%$ & $77.81\%$ & $81.03\%$ & $83.23\%$\\
\hline
\end{tabular}
\end{center}
\end{table}
\subsection{Results of Hyperspectral Data}
Initially, we conduct a $10$-fold cross-validation for the different algorithms on the training set in order to estimate the optimal parameters which can be selected from $\{10^{-2}, 10^{-1}, \allowbreak 10^{0}, 10^{1}, 10^{2}\}$. Table 1 lists classification performances of the different methods with the optimal subspace dimensions obtained by cross-validation using three different classifiers. Correspondingly, the classification maps are given in Figs. 3 and 4 to intuitively highlight the difference.

Overall, PCA performs basically similar performance with the baseline using the three different classifiers on the two datasets. For LPP, due to its sensitivity to noise, it yields a poor performance on the first dataset, while on the relatively high-quality second dataset, LPP steadily outperforms the baseline and PCA. In the supervised algorithms, owing to the limitation of training samples and discriminative power, the classification accuracies of classic LDA is holistically lower than those previously mentioned. With a more powerful discriminative criterion, LFDA obtains more competitive results by locally focusing on discriminative information, which are generally better than those of the baseline, PCA, LPP, and LDA. However, the features learned by LFDA is sensitive to noise and the number of neighbors, resulting in the unstable performance particularly for the different classifiers. For LSDR and LSQMID, they aim to find a linear projection by maximizing the mutual information between input and output from the view of statistics. With fully considering the mutual information, they achieve the good performance on the two given hyperspectral datasets.

Remarkably, the performance of the proposed method (J-Play) is superior to the other methods on the two hyperspectral datasets. This indicates that J-Play is prone to learn a better feature representation and robust against noise. On the other hand, with the increase of $m$, the performance of J-Play steadily increases to the best with around $4$ or $5$ layers for the first dataset and $2$ or $3$ layers for the second one, and then gradually decreases with a slight perturbation since our model is only trained on the training set.
\begin{figure*}[!t]
	  \centering
		\subfigure[Extended Yale-B dataset]{
			\includegraphics[height=5.5cm]{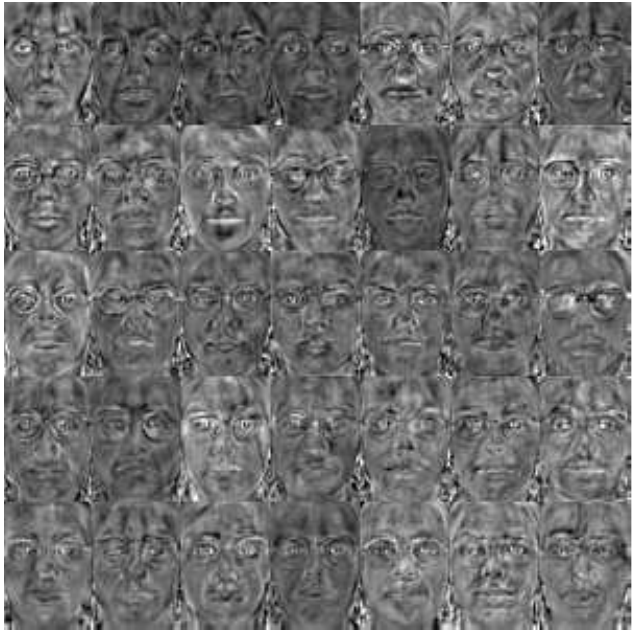}
		}
        \subfigure[AR dataset]{
			\includegraphics[height=5.5cm]{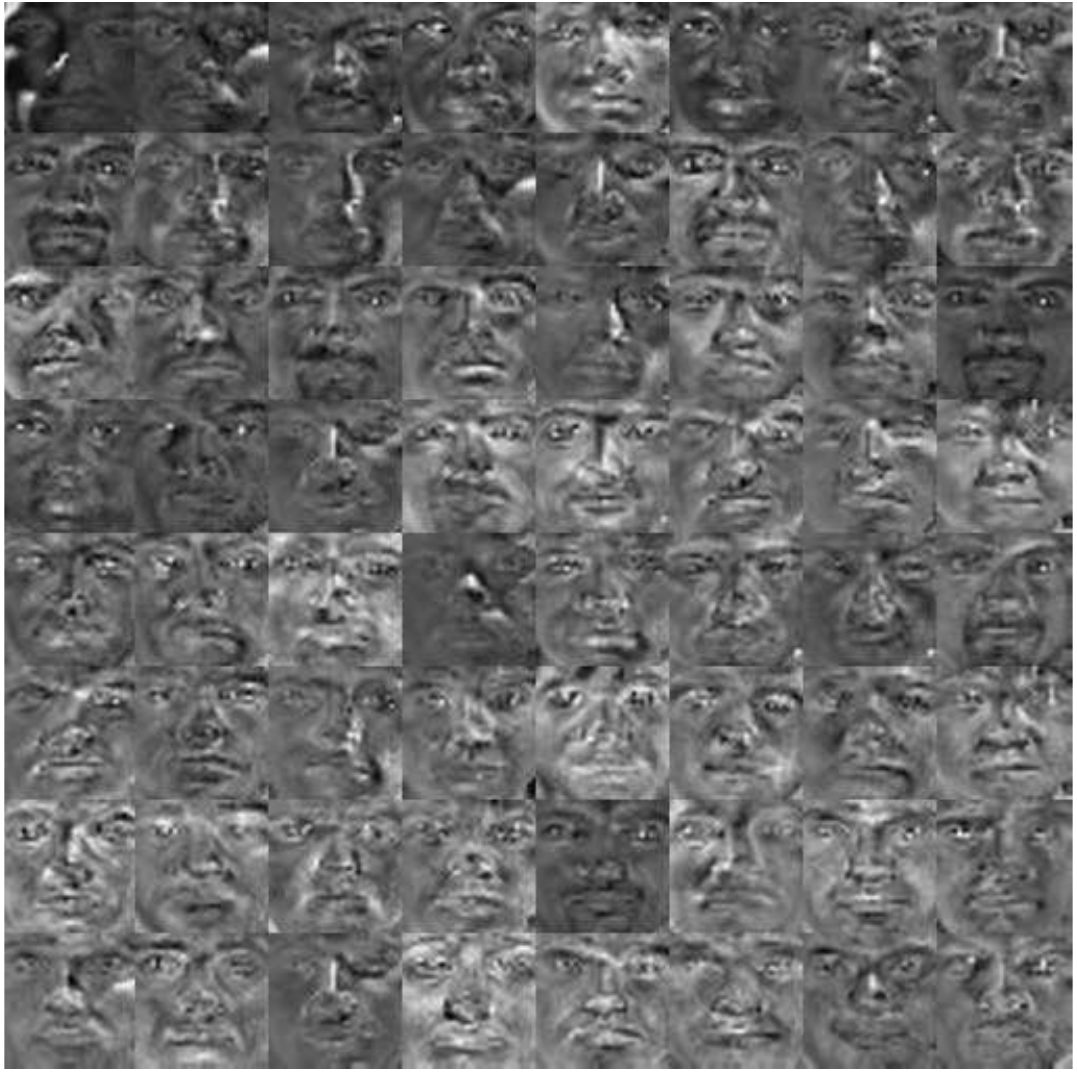}
		}
         \caption{Visualization of partial facial features learned by the proposed J-Play on two face datasets.}
\end{figure*}
\begin{table}[!t]
\begin{center}
\caption{Quantitative performance comparisons on two face datasets. The best results for the different classifiers are shown in red.}
\label{table:headings}
\begin{tabular}{ccccccc}
\hline\noalign{\smallskip}
\multirow{2}{*}{\bf Methods} & \multicolumn{3}{c}{Extended Yale-B dataset} & \multicolumn{3}{c}{AR dataset} \\
\cline{2-7} & NN & KSVM & CCF & NN & KSVM & CCF\\
\hline\noalign{\smallskip}
 Baseline $(1024/2580)$ & $45.77\%$ & $45.87\%$ & $76.99\%$ & $71.71\%$ & $72.29\%$ & $80.29\%$\\
\hline\noalign{\smallskip}
 PCA $(120/80)$ & $41.05\%$ & $81.47\%$ & $83.53\%$ & $68.43\%$ & $80.29\%$ & $81.43\%$\\
\hline\noalign{\smallskip}
 LPP $(170/70)$ & $70.75\%$ & $76.55\%$ & $77.48\%$ & $70.86\%$ & $74.00\%$ & $79.86\%$\\
\hline\noalign{\smallskip}
 LDA $(37/99)$ & $80.88\%$ & $78.37\%$ & $83.68\%$ & $81.43\%$ & $82.29\%$ & $85.38\%$\\
\hline\noalign{\smallskip}
 LFDA $(37/99)$ & $81.02\%$ & $80.88$\% & $83.58\%$ & $71.29\%$ & $75.71\%$ & $80.38\%$\\
\hline\noalign{\smallskip}
 LSDR $(60/80)$ & $71.29\%$ & $76.40\%$ & $78.66\%$ & $75.14\%$ & $79.00\%$ & $80.14\%$\\
\hline\noalign{\smallskip}
 LSQMID $(60/80)$ & $71.48\%$ & $77.09\%$ & $78.37\%$ & $73.29\%$ & $74.29\%$ & $79.29\%$\\
\hline\noalign{\smallskip}
 J-Play$_{1}$ $(170/210)$ & $73.01\%$ & $79.30\%$ & $80.29\%$ & $73.57\%$ & $79.86\%$ & $77.86\%$\\
\hline\noalign{\smallskip}
 J-Play$_{2}$ $(170/210)$ & $81.17\%$ & $84.27\%$ & $85.22\%$ & $82.29\%$ & $86.00\%$ & $84.57\%$\\
\hline\noalign{\smallskip}
 J-Play$_{3}$ $(170/210)$ & $83.43\%$ & $85.50\%$ & $85.76\%$ & $85.43\%$ & $\color{red}88.71\%$ & $87.43\%$\\
\hline\noalign{\smallskip}
 J-Play$_{4}$ $(170/210)$ & $84.07\%$ & $86.09\%$ & $\color{red}86.55\%$ & $85.29\%$ & $87.71\%$ & $87.71\%$\\
\hline\noalign{\smallskip}
 J-Play$_{5}$ $(170/210)$ & $84.56\%$ & $\color{red}86.14\%$ & $86.20\%$ & $85.71\%$ & $87.29\%$ & $\color{red}88.86\%$\\
\hline\noalign{\smallskip}
 J-Play$_{6}$ $(170/210)$ & $85.35\%$ & $85.64\%$ & $86.53\%$ & $85.14\%$ & $87.29\%$ & $88.29\%$\\
\hline\noalign{\smallskip}
 J-Play$_{7}$ $(170/210)$ & $\color{red}85.74\%$ & $85.45\%$ & $86.20\%$ & $\color{red}86.57\%$ & $86.86\%$ & $88.71\%$\\
\hline
\end{tabular}
\end{center}
\end{table}
\subsection{Results of Face Images}
As J-Play is proposed as a general subspace learning framework for multi-label classiciation, we additionally used two popular face datasets to further assess its generalization capability. Similarly, cross-validation on training set is conducted for estimating the optimal parameter combination on the extended Yale-B and AR datasets. Considering the high-dimensional vector-based face images, we first perform the PCA for face images in order to roughly reduce the feature redundancy, whose results are further explored to the dimensionality reduction methods by following the previous work on face recognition  (e.g. LDA (Fisherfaces) \cite{Martínez2001} and LPP (Laplacianfaces) \cite{He2005_2}). Table 2 gives the corresponding OAs using the different methods on the two face datasets respectively.

By comparison, the performance of PCA and LPP is steadily superior to that of baseline, while PCA is even better than LPP. For supervised approaches, LDA performs better than baseline, PCA, LPP and even LFDA, showing an impressive result. Due to the less number of training samples from face datasets, LSDR and LSQMID are limited to effectively estimate the mutual information between the training samples and labels, resulting in the performance degradation compared to the hyperspectral data. The proposed method outperforms other algorithms, which indicates that this method can effectively learn an optimal mapping from original space to label space, further improving the classification accuracy. Likewise, there is a similar trend for the proposed method with the increase of $m$ that J-Play can basically obtain the optimal OAs with around $4$ or $5$ layers and more layers would lead to the performance degradation. We also characterize and visualize each column of the learned projection, as shown in Fig. 5 where those high-level or semantically meaningful features, i.e. face features under the different pose and illumination, can be learned well, making the faces identified easier.
\section{Conclusions}
To effectively find an optimal subspace where the samples can be semantically represented and thereby be better classified or recognized, we proposed a novel linearized subspace learning framework (J-Play) which aims at learning the feature representation from the high-dimensional data in a joint and progressive way. Extensive experiments of multi-label classification are conducted on two types of datasets: hyperspectral images and face images, in comparison with some previously proposed state-of-the-art methods. The promising results using J-Play demonstrate its superiority and effectiveness. In the future, we will further build an unified framework based on J-Play by extending it to semi-supervised learning, transfer learning, or multi-task learning.

\bibliographystyle{splncs04}
\bibliography{egbib}
\end{document}